\documentclass[letterpaper, 10 pt, conference]{ieeeconf}  

\IEEEoverridecommandlockouts                              

\usepackage[english]{babel}
\usepackage{graphicx} 
\usepackage{tikz}
\usepackage{multicol}
\usepackage{placeins}
\usepackage{amsmath} 
\usepackage{amssymb}  
\usepackage{bm}
\makeatletter
\let\NAT@parse\undefined
\makeatother
\usepackage{hyperref} 
\usepackage{pdfpages} 

\title{\LARGE \bf
Motion Planning and Control for Multi Vehicle  \\ Autonomous Racing at High Speeds
}

\author{Ayoub Raji$^{1,2,*}$,  Alexander Liniger$^{3,*}$,  Andrea Giove$^{4,*}$, Alessandro Toschi$^{1}$, Nicola Musiu$^{1}$,\\  Daniele Morra$^{4}$, Micaela Verucchi$^{1}$, Danilo Caporale$^{5}$, Marko Bertogna$^{1}$
\thanks{$^{1}$University of Modena and Reggio Emilia, Italy \newline
        {\tt\small \{ayoub.raji;micaela.verucchi;\newline marko.bertogna\}@unimore.it},\newline
        {\tt\small \{294820;243256\}@studenti.unimore.it},
        }%
\thanks{$^{2}$University of Parma, Italy}
\thanks{$^{3}$Computer Vision Lab, ETH Zurich, Switzerland \newline
        {\tt\small alex.liniger@vision.ee.ethz.ch}}%
\thanks{$^{4}$University of Pisa, Italy \newline
        {\tt\small \{a.giove;d.morra1\}@studenti.unipi.it},
        }%
\thanks{$^{5}$Technology Innovation Institute - \newline Autonomous Robotics Research Center \newline
        {\tt\small danilo.caporale@tii.ae}}%
\thanks{$*$The authors contributed equally.}
}
\newcommand\copyrighttext{%
\footnotesize \copyright 2022 IEEE. Personal use of this material is permitted. Permission from IEEE must be obtained for all other uses, in any current or future media, including reprinting/republishing this material for advertising or promotional purposes, creating new collective works, for resale or
redistribution to servers or lists, or reuse of any copyrighted component of this work in other works.}
\newcommand\copyrightnotice{%
\begin{tikzpicture}[remember picture,overlay]
\node[anchor=north,yshift=-20pt] at (current page.north) {\fbox{\parbox{\dimexpr\textwidth-\fboxsep-\fboxrule\relax}{\copyrighttext}}};
\end{tikzpicture}%
}
\begin{document}
\copyrightnotice
\maketitle
\thispagestyle{empty}
\pagestyle{empty}

\begin{abstract}
This paper presents a multi-layer motion planning and control architecture for autonomous racing, capable of avoiding static obstacles, performing active overtakes, and reaching velocities above 75 m/s.
The used offline global trajectory generation and the online model predictive controller are highly based on optimization and dynamic models of the vehicle, where the tires and camber effects are represented in an extended version of the basic Pacejka Magic Formula. 
The proposed single-track model is identified and validated using multi-body motorsport libraries which allow simulating the vehicle dynamics properly, especially useful when real experimental data are missing.
The fundamental regularization terms and constraints of the controller are tuned to reduce the rate of change of the inputs while assuring an acceptable velocity and path tracking. The motion planning strategy consists of a Frenét-Frame-based planner which considers a forecast of the opponent produced by a Kalman filter. The planner chooses the collision-free path and velocity profile to be tracked on a 3 seconds horizon to realize different goals such as following and overtaking. The proposed solution has been applied on a Dallara AV-21 racecar and tested at oval race tracks achieving lateral accelerations up to 25 m/s\textsuperscript{2}.
\end{abstract}

\section{INTRODUCTION}
\label{introduction_sec}
In the literature, several approaches for motion planning and control have been developed and tested on high-performance autonomous vehicles \cite{survey}. Hierarchical methods which exploit different levels of model complexity at different stages of the motion planner/controller are the current state of the art \cite{laurense_2017, laurense_2022,liniger, vazquez,novi}. The strength of this approach has been shown in \cite{srinivasan}, where a hierarchical method with a Nonlinear Model Predictive Control (NMPC) at its core was able to outperform a top driver on a formula student race car at lateral accelerations of over 20 m/s\textsuperscript{2}.

For the task of multi-vehicle racing, the gap between human expert drivers and autonomous systems is still significant. This is also related to the fundamental challenges that must be solved to tackle this task, which include perception, rule-based interaction with other agents and the infrastructure, motion prediction, generation, and tracking of optimal trajectories for overtakes in unstructured environments. Most related works in this field focus on racing video games, simulations, and RC cars, and very limited work is done on full-scale race cars. In \cite{buyval}, the authors use an NMPC algorithm with a 4-wheel vehicle model with additional states for the nearest obstacle. The solution has been tested in simulation with a control rate of 25 Hz and a maximum speed of 40 m/s. For RC cars several groups have tackled the problem using game-theoretical planners \cite{liniger_game, autorally, stanford_game}, however, these methods focus on one-vs-one racing and do not scale well to full size racetracks. Several algorithms based on Deep Neural Networks (DNN) have also been proposed \cite{weiss,weiss2,pan}. A curriculum reinforcement learning-based method using an off-policy algorithm has been evaluated on Gran Turismo Sports, an arcade racing simulation, outperforming the built-in game AI and reaching similar performance to experienced sim-racing drivers \cite{song}. A different approach is to implement the obstacle avoidance and overtaking tasks in a motion planning module, letting the controller solve only the tracking problem \cite{jung,wang}. In \cite{stahl}, the authors present a multi-layered graph-based planning architecture in which the trajectory is chosen considering a cost function representing the feasibility of the vehicle to follow the path segments of the graph built offline. The method has been tested in a real-world overtaking maneuver at low speeds in a simplified adversarial context.
\begin{figure}[t]
	\centering
	\includegraphics[width=1\columnwidth]{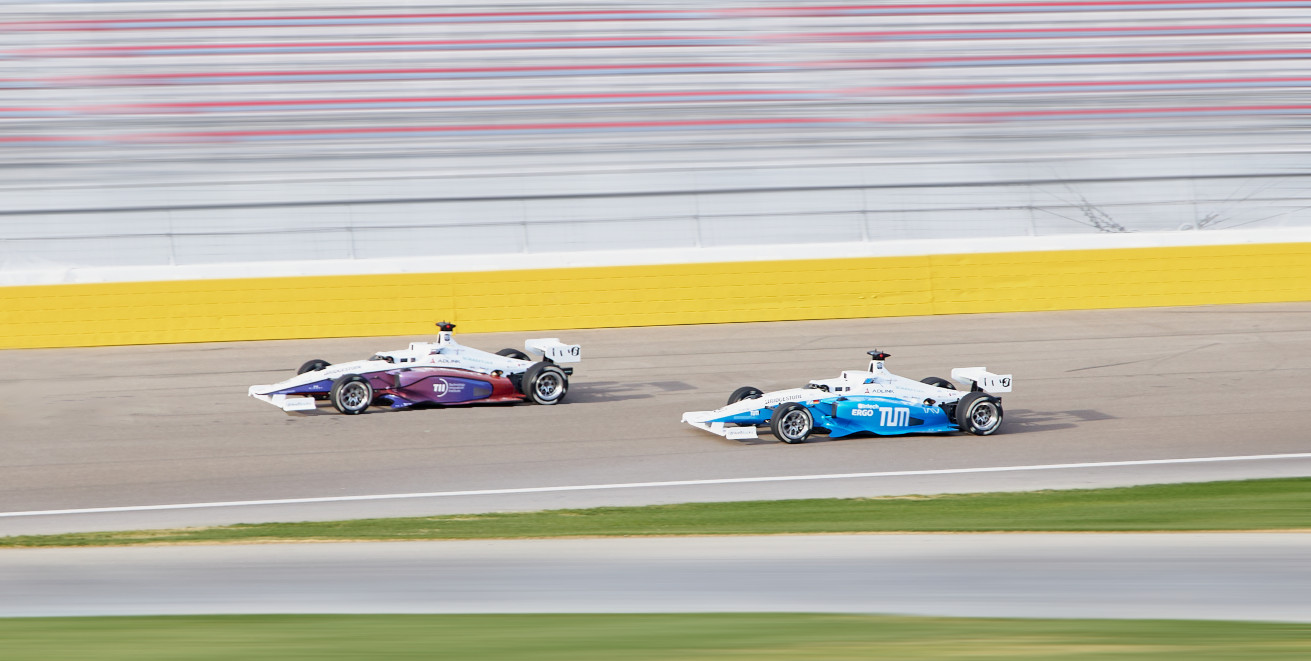}
    \caption{TII EuroRacing overtaking TUM Autonomous Motorsport during semifinal of the Autonomous Challenge at CES, Las Vegas Motor Speedway. \copyright Yev Z Photography}
	\label{fig:TIIvsTUM}
\end{figure}

In this article, we present a framework for planning and control in head-to-head autonomous racing conditions evaluated during the Indy Autonomous Challenge (IAC{\footnote{\href{https://www.indyautonomouschallenge.com/}{https://www.indyautonomouschallenge.com/}}) events at the Indianapolis Motor Speedway (IMS) and Las Vegas Motor Speedway (LVMS) on a full-scale open-wheel racecar. It has been reached a top speed of 75 m/s in a single vehicle scenario, and a speed of 63 m/s during an overtaking maneuver. The authors participated in the competition as part of the TII EuroRacing team (TII-ER). 

In Section \ref{model_sec} we describe the vehicle model used for the optimization-based problems and introduce our model identification and validation approach. In Section \ref{planning_sec} the lap time optimization strategy is presented before describing the motion forecasting and the Frenét-Frame-based method for the local planning. Constraints, cost function, and tuning strategies applied on the controller are reported and explained in Section \ref{mpc_sec}. The experimental results are described in section \ref{results_sec}, while final conclusions and future works are discussed in Section \ref{conclusion_sec}}.

\section{VEHICLE MODEL}
\label{model_sec}
The vehicle considered in this paper is a Dallara AV-21, shown in Figure \ref{fig:TIIvsTUM}, based on the Indy Lights chassis IL-15 with a 390hp engine. The suspensions and aerodynamics are adjusted for oval racing with an asymmetrical setup to exploit highly banked tracks.

\subsection{Curvilinear Single Track Model}

\begin{figure}[htb!]
	\centering
	\includegraphics[width=1.0\columnwidth]{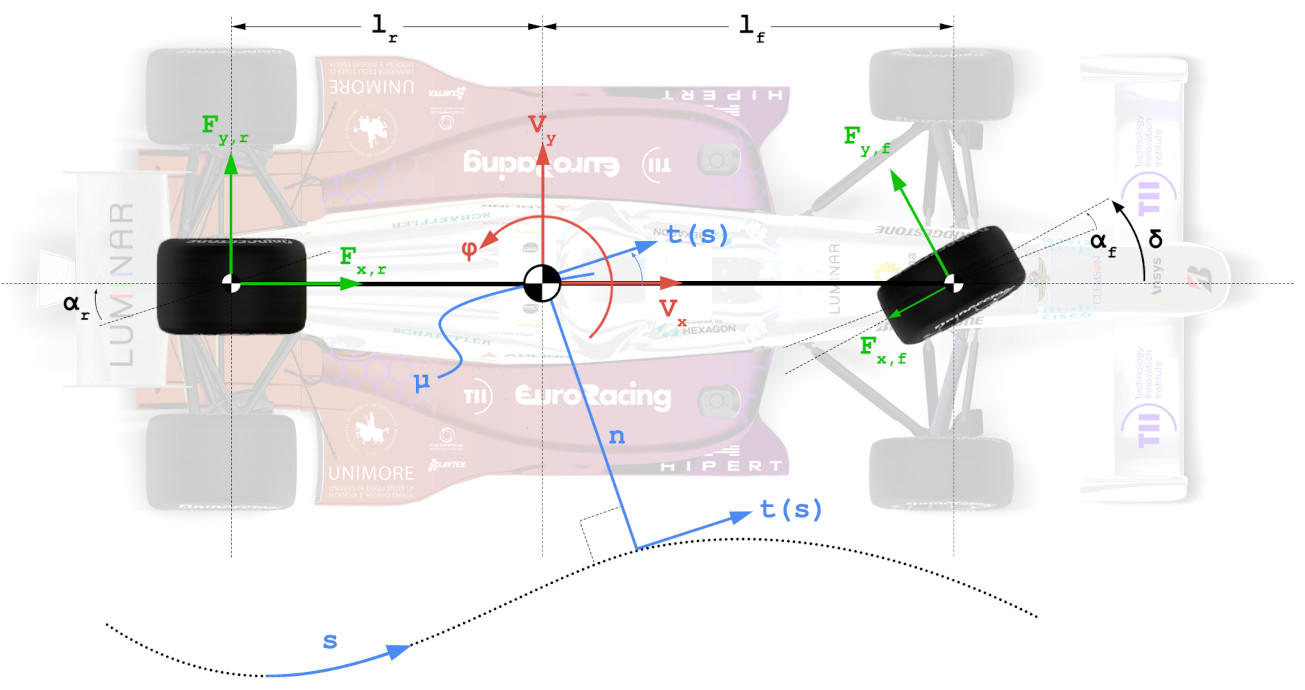}
    \caption{Dynamic single track vehicle model on curvilinear coordinates.}
	\label{fig:simulink_mod}
\end{figure}

A single track dynamic model, shown in Figure \ref{fig:simulink_mod}, is used for the offline trajectory optimization problem and the NMPC. As in \cite{vazquez}, we use curvilinear/Frenét coordinates to describe the state. Therefore, global position and heading are not directly considered, but transformed to a state relative to the reference path.

\subsubsection{Equations of Motion}

The vehicle state is given by $x=[s; n; \mu; v_x; v_y; r; \delta; T; B]$ and the input as $u = [\Delta\delta; \Delta T; \Delta B]$, where $s$, $n$ and $\mu$ are the progress along the path, the orthogonal deviation from the path and the local heading. Longitudinal $v_x$ and lateral $v_y$ velocities are considered as well as the yaw rate $r$. $\delta$, $T$ and $B$ are the steering angle, throttle command and brake command, which are included in the state. The control commands $\Delta\delta$, $\Delta T$ and $\Delta B$ are the derivatives of the inputs.
Thus, the equations of motion are
\begin{small}
\begin{align*}
    \label{eq:model:motion}
        \dot{s} =& \, \frac{v_x \, \cos(\mu) - v_y \, \sin(\mu)}{1 - n \, \kappa(s)} \,,\nonumber\\
        \dot{n} =& \, v_x \, \sin(\mu) + v_y \, \cos(\mu)\,,\nonumber\\
        \dot{\mu} =& \, r - \kappa(s) \dot{s}\,,\nonumber\\
        \dot{v}_x =& \, \frac{\scalebox{0.8}{1}}{\scalebox{0.8}{\textit{m}}}\big(F_{x_r} - F_{d}- F_{y_f} \sin(\delta) + F_{x_f} \cos(\delta) - F_{b_x} + m v_y r\big)\,,\nonumber\\
        \dot{v}_y =& \, \frac{\scalebox{0.8}{1}}{\scalebox{0.8}{\textit{m}}}\big(F_{y_r} + F_{y_f} \cos(\delta) + F_{x_f} \sin(\delta) - F_{b_y} - m v_x r\big)\,,\nonumber\\
        \dot{r} =& \, \frac{\scalebox{0.8}{1}}{\scalebox{0.8}{\textit{I}}_z}\Big(l_f \big(F_{y_f} \cos(\delta) + F_{x_f} \sin(\delta)\big) - l_r F_{y_r}\Big)\,,\nonumber\\
        \dot{\delta} =& \, \Delta\delta \,,\nonumber\\
        \dot{T} =& \, \Delta T \,,\nonumber\\
        \dot{B} =& \, \Delta B \,,
\end{align*}
\end{small}
where $\kappa(s)$ is the curvature at the progress $s$, $l_f$ and $l_r$ are the distances from the center of gravity to the front and rear wheels, $m$ is the mass and $I_z$ the moment of inertia. $F_{y_f}$, $F_{y_r}$ are the lateral tire forces at the front and rear wheels. $F_{x_f}$, $F_{x_r}$ are the longitudinal forces at front and rear axles.
$F_{b_x}$ and $F_{b_y}$ model the forces on the x and y-axis due to the road bank angle $\theta$, and are given by $F_{b_x} = \, m g \, \sin(\theta) \, \sin(\mu)$ and $F_{b_y} = \, m g \, \sin(\theta) \, \cos(\mu)$. $F_{d}$ represents the aerodynamic effects considering the air density $\rho$, the frontal area $S$ and the drag coefficient $C_d$,
\begin{align*}
    \begin{split}
        F_{d} =& \, 0.5 \, \rho \, S \, C_d \, v_{x}^{2}\,.
    \end{split}
\end{align*}

\subsubsection{Tire Model}

The tires effects are modeled using a simplified Pacejka Magic Formula \cite{pacejka} with a combined slip correction. Beyond the usual macro-parameters $B$, $C$, $D$ and $E$, the lateral force offsets $Sv_{yi}$, $i \in [f, r]$, have been included resulting in
\label{eq:model:pacejka}
\begin{align}
    \begin{split}
        F_{y_f,\text{lat}} =& \, Sv_{y_f} + D_{f} \, \sin\Big(C_{f} \tan^{-1}(B_{f} \, \alpha_{y_f}) +\\
                & - E_{f} \big(B_{f} \, \alpha_{y_f} - \tan^{-1}(B_{f} \, \alpha_{y_f})\big)\Big)\,,\\
        F_{y_r,\text{lat}} =& \, Sv_{y_r} + D_{r} \sin\Big(C_{r} \tan^{-1}(B_{r} \, \alpha_{y_r}) +\\
                & - E_{r} \big(B_{r} \, \alpha_{y_r} - \tan^{-1}(B_{r} \, \alpha_{y_r})\big)\Big)\,,
    \end{split}
\end{align}
where $\alpha_{y_i} = \alpha_i + Sh_{y_i}$ is the resulting slip angle obtained by applying a shift $Sh_{y_i}$ to the front slip angle $\alpha_f$ and the rear slip angle $\alpha_r$ 
\begin{align*}
    \begin{split}
        \alpha_f =& \, \tan^{-1}\left(\frac{v_y + l_f  \, r}{v_x}\right) - \delta\,,\\
        \alpha_r =& \, \tan^{-1}\left(\frac{v_y - l_r \, r}{v_x}\right)\,.
    \end{split}
\end{align*}
$Sh_{y_i}$ and $Sh_{v_i}$ are calculated using Pacejka micro-parameters \cite{pacejka} related to horizontal shifts and variation of the lateral force shift considering
the change in the tire load with respect to the reference vertical load and the camber angle.

To consider the combined slip, we propose a combined slip weighting factor. Thus, the pure lateral forces $F_{y_f,lat}$ and $F_{y_r,lat}$ are weighted with $G_{y_i}$, such that we get the final forces, $F_{y_f} = G_{y_f} F_{y_f,lat}$ and $F_{y_r} = G_{y_r} F_{y_r,lat}$, with $G_{y_i}$ given by
\begin{align*}
    G_{y_i} = \cos\big( \arcsin ( F_{x_i} / F_{\max_i})\big) \,,
\end{align*}
where $F_{\max_i} = D_i \epsilon_i$ and $\epsilon_i$ is an ellipse shape parameter. Note that we clip the force fraction $F_{x_i} / F_{\max_i}$ at 0.98 to avoid singularity issues. 

\subsubsection{Longitudinal Forces}

The front axle longitudinal forces are modeled as
\begin{equation*}
    F_{x_f} = - C_{b_f} B - C_{ro}\,,
\end{equation*}
where $C_{ro}$ is the rolling resistance. The braking force is represented as $C_{b_f} B$, with $C_{b_f}$ the maximum brake pedal pressure and $B \in [0, 1]$.

Considering a rear wheels drive powertrain, the rear longitudinal force is modeled as
\begin{equation} \label{eq:powertrain}
    F_{x_r} = C_m T - C_{br} B - C_{ro}\,,
\end{equation}
where $C_m$ is a linear engine coefficient and $T \in [0, 1]$. The turbo-charged combustion engine used on the research vehicle produces a force which is not linear in the whole usable regions since it depends on the engine rpm and gear. In order to represent this behavior on \eqref{eq:powertrain}, a scale factor $k_{s}$ varying on the speed has been applied to the upper bound constraint of the throttle command, thus $T \in [0, k_{s}]$. 

Gear shifting effects are neglected as well as the gear command which is controlled separately and sent to the low-level controller when reaching the desired engine rpm.

\subsection{Model Identification}

Due to the lack of a steering wheel on the research vehicle, the traditional maneuvers used to collect data for vehicle model identification were not practical \cite{farroni,denhartog}.
We relied on information provided by the IAC organizers and tire and vehicle manufacturers, initially limiting our model to a static identification.

\subsubsection{Multi Body Simulation}

Dymola \cite{dymola}, a physical modeling and simulation tool, has been used to model the AV-21 vehicle dynamics with the VeSyMA - Motorsports Library \cite{dempsey}. The library provides solutions to model open-wheel race-cars components such as suspensions, aerodynamics, tires, and the powertrain.
A highly detailed multi body simulation of the vehicle has been developed using the available information on the mechanical components such as the static parameters of the Indy Lights chassis and the engine map retrieved from a test bench. Unknown components or possible setups choices for the suspension have been estimated from IL-15 and IndyCar oval configurations, in particular the camber, caster and toe.
The IMS and LVMS tracks have been modeled on Dymola by estimating the banking angle from sim-racing games. The resulting tracks have later been validated using the data from the on-board LiDAR sensors.

\subsubsection{Tire Model Identification}

The tire maker provided a Magic Formula 6.2 model obtained using a test rig. However, the model could not be used to reproduce accurately enough the real tire behavior, which is highly affected by the tire-road grip, wear and suspension setup. A common strategy is to use the provided set of coefficients as a starting point and run an identification procedure to find parameters that better match the experimental data gathered on track.

In our work, the approach presented in \cite{farroni} has been applied using data obtained by simulating ramp steer maneuvers at different speeds and road conditions in our Dymola simulator.

\subsubsection{Validation}

\begin{figure}
	\centering
	\includegraphics[width=1.0\columnwidth]{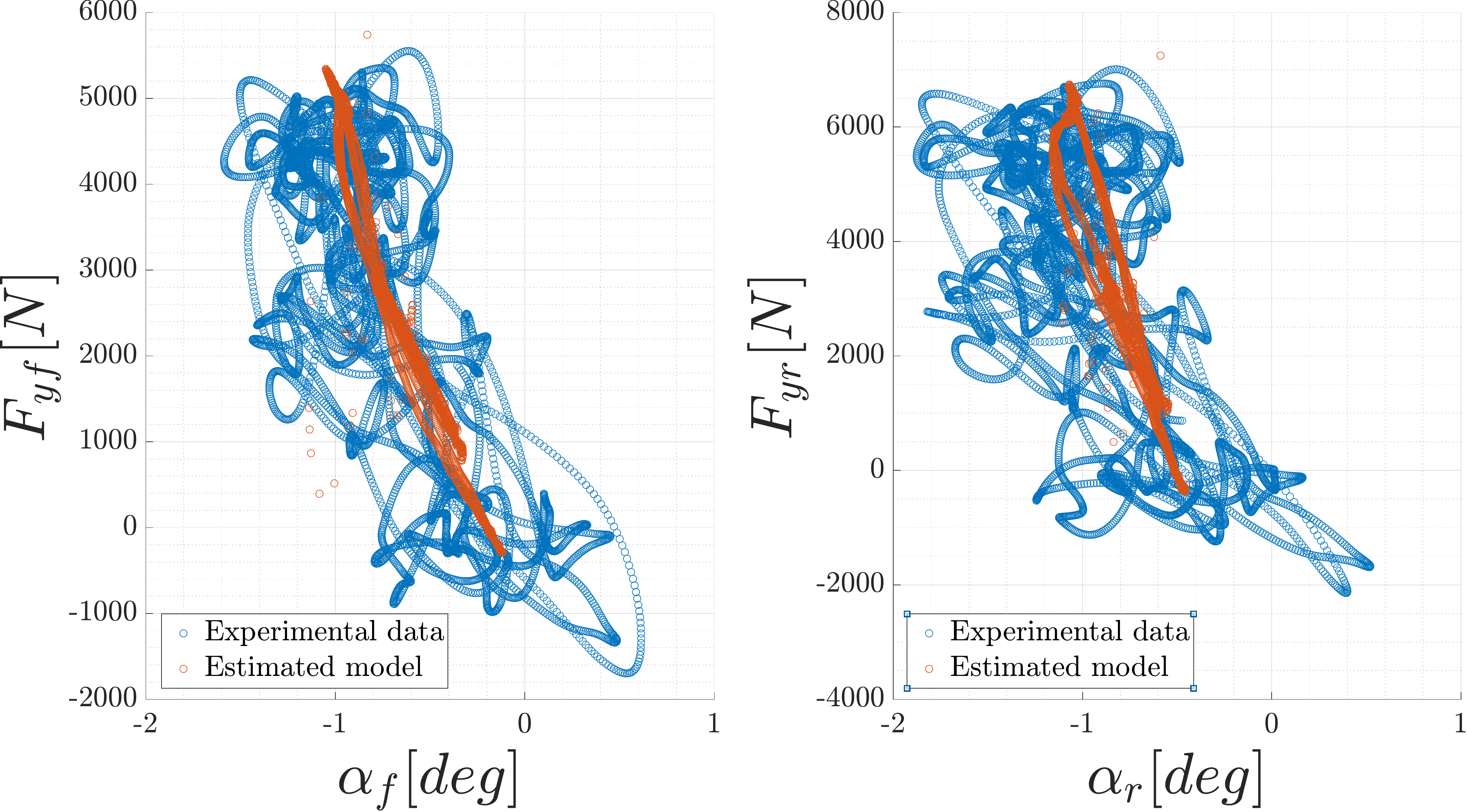}
    \caption{The estimated tire model, lateral force on the front (left) and rear (right) axles, over the real data gathered at LVMS at a speed of 62 m/s.}
	\label{fig:lvms_62ms}
\end{figure}

\begin{figure}
	\centering
	\includegraphics[width=1.0\columnwidth]{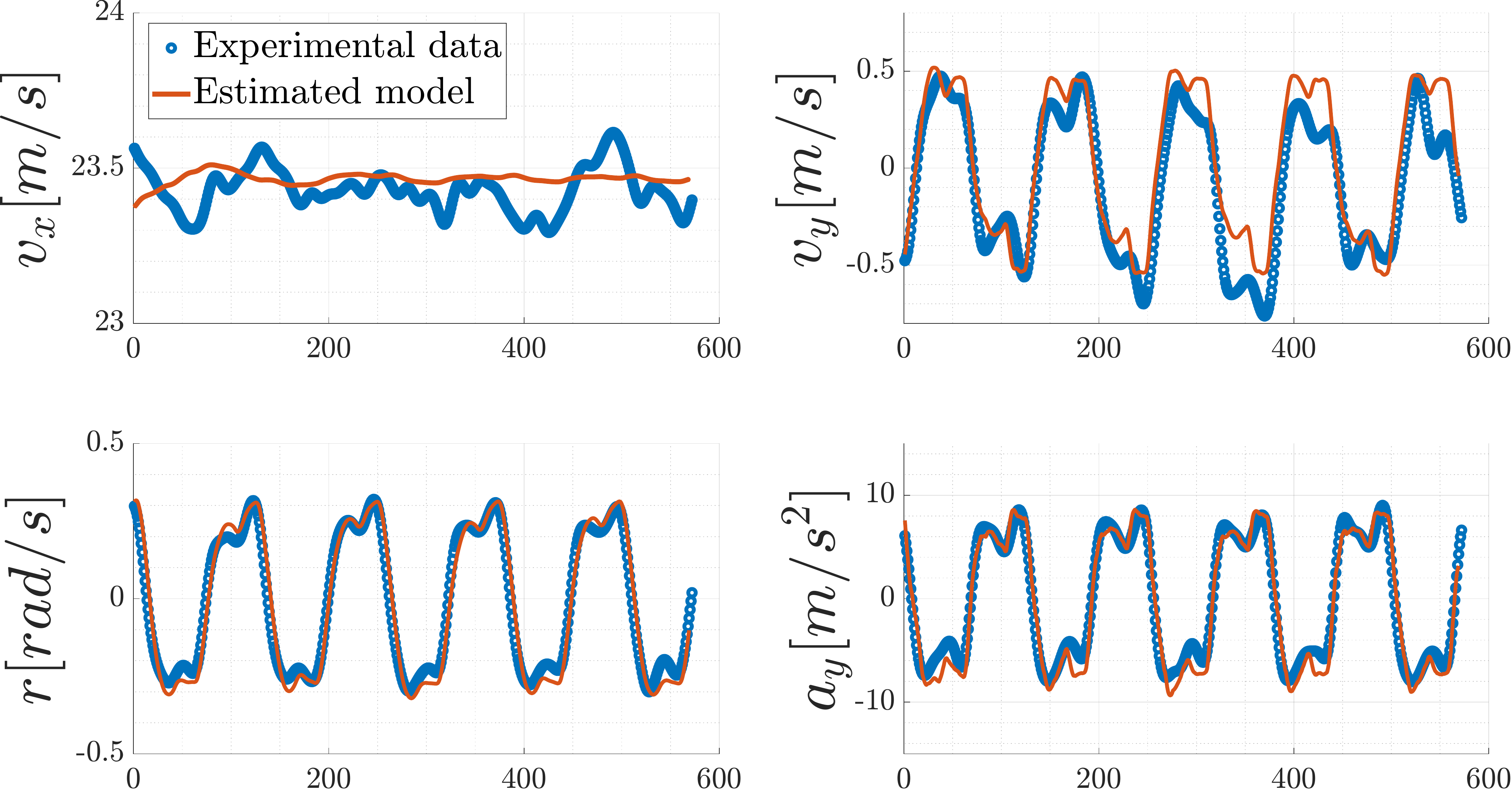}
    \caption{Comparison of real and simulated data of the warm-up maneuver.}
	\label{fig:ims_warmup}
\end{figure}

First experimental data on the real vehicle have been gathered using a simple Pure Pursuit path tracking algorithm \cite{coutler} at a maximum speed of 45 m/s at IMS and performing a light warm-up maneuver at 25 m/s in the long straights of the track. The warm-up maneuver consists of a series of $\pm$80deg steering wheel angle reference jumps on top of the lateral controller. 
An optical sensor has been mounted to get accurate measurements of speeds and angles in addition to the data obtained from the GNSS RTK-corrected system available on the Dallara AV-21. The tire model fitted on real data is depicted in Figure \ref{fig:lvms_62ms}. In Figure \ref{fig:ims_warmup}, a comparison of the real and simulated data of the warm-up maneuver is shown. 
The first set of simplified Pacejka coefficients estimated from Dymola has been used during the tests and races at IMS and LVMS in the MPC described in Section \ref{mpc_sec}.

\section{MOTION PLANNING DESIGN}
\label{planning_sec}

\subsection{Offline Global Trajectory Generation} 

A global path is generated as the main reference for the local planner. The resulting global path should consider the dynamic model, constraints on the inputs and tires, but should also give the possibility to incorporate rules related to track limits, such as keeping an inner or outer line.

Following \cite{vazquez}, the solution is found by solving an optimal control problem using the dynamics transformed in the spatial domain $f_s\big(x(s),u(s)\big)$, with the progress $s$ as running variable. The continuous space model is discretized with a discretization distance $\Delta_s$ resulting in $x_{k+1} = f_s^d(x_k, u_k) = x_k + \Delta_s f_s(x_k, u_k)$.

The cost function maximizes the progress rate $\dot{s}$ including a regularization term $B(x_k) = q_B \alpha_r^2$ which penalizes the rear slip angle, and a regularizer on the input rates $u^T Ru$ where $R$ is a diagonal weight matrix. In summary, the overall cost function is defined as
\begin{equation}
\label{eq:global:cost}
   J_{opt}(x_k, u_k) = -\dot{s}_k + u^T Ru + B(x_k) \,.
\end{equation}
Combining the cost, model, and constraints the optimization problem is formulated as
\begin{align*}
    \begin{split}
        \min_{X, U} &\ \ \sum_{k=0}^{N} = J_{opt}(x_k, u_k) \\
        s.t. &\ \ x_{k+1} = f_s^d(x_k, u_k)\,, \\
        &\ \ f_s^d(x_N, u_N) = x_0\,, \\
        &\ \ x_k \in X_{track} \quad x_k \in X_{ellipse}\,,\\
        &\ \ a_k \in \bm A, \, u_k \in \bm U,  \ k = 0,\dots,N, 
    \end{split}
\end{align*}
where $X=[x_0, ..., x_N]$, and $U=[u_0, ..., u_N]$. $X_{ellipse}$ represents velocity dependent friction ellipse constraints similar to \cite{srinivasan}, and $X_{track}$ represents a track constraint on the lateral deviation $n$ ensuring that the trajectory stays on the track, considering additional side margin distances $n_{\text{left}}$, $n_{\text{right}}$ to the half length $L_c$ and half width $W_c$ of the car, and the left and right track width $N_{L/R}$ at a progress $s$,
\begin{align}
    \begin{split}
   n + L_c \sin|\mu| + W_c \cos\mu \leq N_L(s) -n_{\text{left}} \,, \\
   - n + L_c \sin|\mu| + W_c \cos\mu \leq N_R(s) + n_{\text{right}}\,.
    \end{split}
\end{align}
The physical inputs $a = [\delta; T; B]$ and their rate of change $u$ are constrained using box constraints $\bm{A}$ and $\bm{U}$. The problem is formulated in JuMP \cite{jump} and solved using IPOPT \cite{ipopt}.

\subsection{Motion Forecasting}

The motion forecasting module receives the position of the moving obstacles from the perception module, which processes the raw information of the sensors and keeps track of the obstacles in time. The module assigns to each obstacle a unique identifier $i$, a position in a Cartesian frame $x_i, y_i$, and a covariance matrix of the position $\Sigma_{xy,i}$.

Starting from the position of the $i$-th obstacle in a Cartesian frame $x_i(k), y_i(k)$ at step $k$, the position of the obstacle in the Frenét frame $s_i(k), n_i(k)$ is computed. 
Then, we define the model of the obstacle as
\begin{align}
  \label{eq:forecasting:model_dot_s}
  \dot s_i(k + 1) &= \dot s_i(k)\,, \\
  \label{eq:forecasting:model_d}
  n_i(k + 1) &= n_i(k)\,.
\end{align}
Equation \eqref{eq:forecasting:model_dot_s} states that the longitudinal speed of the obstacle is constant, whereas equation \eqref{eq:forecasting:model_d} indicates that the lateral displacement to the reference path is constant.

This simple model exploits the fact that the only obstacles in the track are other cars that will follow a racing line similar to the one that the ego car is following, and that the speed on an oval race track is almost constant. 
This insight is perfectly represented by modeling objects in a Frenét frame that uses the race line as the reference path, combined with our motion model.

From the equations \eqref{eq:forecasting:model_dot_s}, \eqref{eq:forecasting:model_d}, the following state space model is derived
\begin{align}
  \mathcal{X}_i(k+1) &=
  \left[
    \begin{array}{c}
      s_i(k + 1) \\
      \dot s_i(k + 1) \\
      n_i(k + 1)
    \end{array}
  \right ]
  =
  \left [
    \begin{array}{cccc}
      1  & T_s & 0  \\
      0  & 1   & 0  \\
      0  & 0   & 1 
    \end{array}
  \right ]
  \mathcal{X}_i(k)\,, \\
  \mathcal{Y}_i(k) &= 
  \left [
    \begin{array}{cccc}
      1  & 0 & 0 \\
      0  & 0 & 1
    \end{array}
  \right] 
  \mathcal{X}_i(k)\,,
\end{align} 
where $T_s$ is the sampling period of the filter, $\mathcal{X}_i(k)$ is the state of the model and $\mathcal{Y}_i(k)$ is the output.

Thus, at each time step for every obstacle, the Frenét frame measurements $\hat s_i(k), \hat n_i(k)$ are computed from $\hat x_i(k), \hat y_i(k)$. Using these measurements, the Kalman filter is updated with a prediction step, followed by a correction phase in which $\Sigma_{sn}$, the covariance matrix of the position converted in the Frenét frame, is used. The future trajectory of the obstacle $\mathcal{O}_i(k+1|k), \dots, \mathcal{O}_i(k+m|k)$ is predicted by applying $m$ consecutive prediction steps.

\subsection{Frenét-Frame-based Planner}

The planner module implemented is an extended version of \cite{werling} adding further considerations to the moving obstacles' collision check and racing scenarios. 

\subsubsection{Trajectories Generation}

Given a race line computed offline, a Frenét frame is defined and used to generate multiple trajectories, which for example merge to the reference line, follow a vehicle or perform an overtake. 
Each single trajectory is defined as a combination of a lateral movement $n(t)$ and a longitudinal movement $s(t)$ at time $t$ with respect to the reference path.

Starting from the lateral movements, let $N_0 = [n_0, \dot n_0, \ddot n_0]$ be the start state and $N_1 = [n_1, \dot n_1, \ddot n_1]$ be the end state.
As we want to move parallel to the reference line, we generate the set of lateral movements by changing $n_1$ in an interval $[n_{\text{min}}, n_{\text{max}}]$ and set $\dot n_1 = \ddot n_1 = 0$.
Given $N_0$, $N_1$ and the time interval $T$ between them, a quadratic polynomial is fully defined, and its coefficients can be calculated.
For each lateral movement, we then assign a cost based on the following cost function:
\begin{equation*}
  C_{\text{lat}} = k_j J_t\big(n(t)\big) + k_t T + k_d n_1^2\,.
\end{equation*}
This cost function penalizes the solutions with slow convergence to the reference, i.e. the ones that at the end of the trajectory are off from the reference path $n = 0$.
Unlike what is proposed in \cite{werling}, we decided to keep $T$ constant for all the trajectories in the set in order to provide to the controller a trajectory with a fixed time horizon.

A similar approach has been used for the longitudinal movement generating trajectories that bring the car to a desired velocity $\dot s_n$, while minimizing the jerk.
As shown in \cite{werling}, quartic polynomials can be found to minimize the cost function
\begin{equation}
\label{eq:planner:c_long}
  C_{\text{long}} = k_j J_t\big(s(t)\big) + k_t T + k_{\dot s} (\dot s_1 - \dot s_n)^2
\end{equation}
for a given start state $S_0 = [s_0, \dot s_0, \ddot s_0]$ at $t_0$ and $[\dot s_1, \ddot s_1]$ of the end state $S_1$ at some $t_1 = t_0 + T$.
This means that we can generate a set of optimal longitudinal trajectories by varying the end constraints $\dot s_1 = \dot s_n + \Delta \dot s$ and $T$.

The set of lateral movements $\mathcal{T}_ {\text{lat}}$ and longitudinal movements $\mathcal{T}_ {\text{lon}}$ are then combined, resulting in a set $\mathcal{T} = \mathcal{T}_ {\text{lat}} \times \mathcal{T}_ {\text{lon}}$ of complete trajectories.

\subsubsection{Trajectory Selection}

All the trajectories $\tau_i \in \mathcal{T}$ are checked to evaluate whether they exceed the track boundaries or collide with an obstacle.
We decided to perform these checks in the Frenét frame, to avoid converting the trajectories to a Cartesian frame. Furthermore, rather than doing the checks on the polynomials, we sampled each trajectory in a finite number of points.
The sampling is done by fixing a time interval $\Delta t$ and evaluating the trajectory in $\Delta t, 2 \Delta t, \dots, M \Delta t$.
Thus, the trajectory is converted to a set of points, where each point is associated with a time instant:
\begin{equation*}
  \tau_i \rightarrow \bar \tau_i = \big\{ \{ t_k, (s_{\tau_i,k}, n_{\tau_i, k} )\} , \quad k = 1, \dots, M  \big\}\,.
\end{equation*}
Given the track width in every point of the reference path, it is trivial to check if a trajectory $\bar \tau_i$ goes out of the track boundaries.

From the Motion Forecasting module, an obstacle $\mathcal{O}_j$ is defined as a set of points. Each point is associated with a time instant:
\begin{equation*}
  \mathcal{O}_j = \big\{ \{t_k, (s_{o_j,k}, \; n_{o_j, k} ) \} , \quad k = 1, \dots, M \big\}\,.
\end{equation*}

To account for the safety margins, a rectangle is built around every point of the predicted trajectory of the obstacle. A trajectory $\bar \tau_i$ collides with an obstacle $\mathcal{O}_j$ if $\exists k \in \{1, \dots, M \}$ such that $(s_{\tau_i,k}, n_{\tau_i, k} )$ is inside the rectangle built around $( s_{o_i,k}, n_{o_i, k} )$.
The described collision check treats the obstacle as a hard constraint for the planning algorithm. This approach could lead to undesired behavior in the scenarios in which an obstacle is blocking the race line, because the best trajectory will be the nearest one to the obstacle that does not collide with the obstacle itself. Following such a trajectory could bring the car to the edge of the obstacle safety margins. This can be critical if we consider the possible noise in the detection module.
To overcome this issue, the collision check method is improved by adding a soft constraint.
For each trajectory $\tau_i$, a collision coefficient $\gamma_i \in [0, 1]$ is computed, where $\gamma_i = 0$ indicates that the trajectory is not colliding with any obstacle, whereas $\gamma_i = 1$ indicates that the trajectory is violating the safety margins (hard constraint).
Given this change, the cost becomes
\begin{equation*}
  C_{\text{tot},i} = k_{\text{lat}} C_{\text{lat},i} + k_{\text{lon}} C_{\text{lon},i} + k_{\text{soft}} \gamma_i^2 \,,
\end{equation*}
with $k_{\text{lat}}, k_{\text{lon}}, k_{\text{soft}} > 0$.

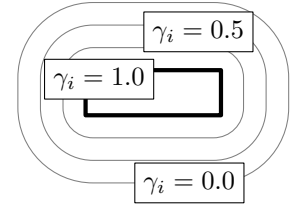
\begin{figure}[htb!]
  \centering
  \begin{tikzpicture}[line cap=round, line join=round, x=1cm, y=1cm, scale=0.60]
  \draw[ultra thick] (0,0) rectangle (3, 1);
  

  \draw[gray,rounded corners=1.0cm] (-1.5,2.5) -- (4.5,2.5) -- (4.5, -1.5) -- (-1.5,-1.5) -- cycle;
  \draw[gray,rounded corners=0.7cm] (-1,-1) -- (-1,2) -- (4,2) -- (4, -1) -- cycle;
  \draw[gray,rounded corners=0.4cm] (-0.5,-0.5) -- (-0.5,1.5) -- (3.5,1.5) -- (3.5, -0.5) -- cycle;

  \node at (0.3,0.8) [draw,minimum width=1cm, fill=white] {$\gamma_i = 1.0$};
  \node at (2.5,1.85) [draw,minimum width=1cm, fill=white] {$\gamma_i = 0.5$};
  \node at (2.3,-1.5) [draw,minimum width=1cm, fill=white] {$\gamma_i = 0.0$};
\end{tikzpicture}
  \caption{Graphical representation of the collision coefficient as defined in \eqref{eq:collision_coef}. The black rectangle indicates the safety margin from the obstacle (hard constraint).}
  \label{fig:soft_constraint}
\end{figure}

To compute $\gamma_i$ we decided to exploit the Euclidean distance from the safety margin.
For every trajectory $\tau_i$ the minimum distance $n_i$ from the safety margin is computed.
Then, $\gamma_i$ is defined as
\begin{equation}\label{eq:collision_coef}
  \gamma_i = \max \left \{1 - \frac{n_i}{\Delta_{soft}}, 0 \right \}\,,
\end{equation} 
where $\Delta_{soft} > 0$ is a parameter to enlarge or reduce the effect of the soft constraint.
In Figure \ref{fig:soft_constraint} a graphical representation of the soft constraint is given.

The final step of our planner is to select the trajectory with the minimal cost which does stay inside the track margins.

\subsubsection{Following Mode}

A following mode is implemented in the planner to keep a desired distance to the opponent when it is not allowed to perform an overtake by the rules of the competition.
Differently from \cite{werling}, given the position $\bm{P}_{\text{opp}} = [s_{\text{opp}}, n_{\text{opp}}]$ and speed $\dot s_{\text{opp}}$ of the car to keep the distance from and the desired distance $\Delta_{\text{des}}$, the desired speed of the ego car $\dot s_{\text{des}}$ is regulated by a simple proportional controller:
\begin{equation*} \label{eq:planner:distance_keeping}
  \dot s_{\text{des}} = \dot s_{\text{opp}} - k \left ( \Delta_{\text{des}} - \Delta_{\text{gap}} \right )
\end{equation*}
where $\Delta_{\text{gap}} > 0$ is the current distance between the ego and the opponent car, and $k > 0$ is the gain of the controller.
The computed desired speed is then used in \eqref{eq:planner:c_long} for the longitudinal movement generation.


\section{MODEL PREDICTIVE CONTROL DESIGN}
\label{mpc_sec}
\subsection{MPC Problem}

In the control problem, the model is discretized in time $f_t^d(x_t, u_t)$ using a fourth-order Runge Kutta method.
As in \eqref{eq:global:cost}, the MPC cost function combines the progress optimization with the regularization terms in order to penalize the rate of change of the physical inputs and rear slip angle:
\begin{equation*}
   J_{MPC}(x_t, u_t) = -\dot{s}_t + q_n n_t^2 + q_\mu \mu_t^2 + q_v | s_{v,t} | + u^T Ru + B(x_t)\,.
\end{equation*}
In addition to \eqref{eq:global:cost}, the cost function includes path following weights $q_n$ and $q_\mu$, as well as a velocity tracking weight $q_v$ on the slack variable $s_{v,t}$ of the upper velocity constraint. Note that the reference path is given by the Frenét-planner. 

The MPC problem is formulated as
\begin{align*}
    \begin{split}
        \min_{X, U} &\ \ \sum_{t=0}^{T} = J_{MPC}(x_t, u_t) \\
        s.t. &\ \ x_0 = \hat{x} \,,\\
        &\ \ x_{t+1} = f_t^d(x_t, u_t)\,, \\
        &\ \ x_t \in X_{track} \quad x_t \in X_{ellipse}\,,\\
        &\ \ v_{x,t} \leq \bar{v} - s_{v,t} \quad s_{v,t} \geq 0\,,\\
        &\ \ a_t \in \bm A, \, u_t \in \bm U, t = 0,\dots,T.
    \end{split}
\end{align*}
where $\hat{x}$ is the current curvilinear state, $\bar v$ the upper velocity bound and T is the prediction horizon. The main difference to \cite{vazquez} is the more complex model and the integration with the Frenét-planner.

The optimization problem is solved using a custom sequential quadratic programming framework, which uses HPIPM \cite{frison}, a high-performance quadratic programming framework for MPC, and CppADCodeGen, a code generation automatic differentiation library. 

\subsection{Tuning}

The regularization weights and constraints have been chosen in order to manage the trade-off between low path tracking error and input commands smoothness. Due to the uncertainty of the actuation performance and the model mismatch at high speeds, it has been decided to set the regularization term related to the steering wheel rate of change one order of magnitude higher than the value used in simulation and in other autonomous racing platforms in which the controller has been tested.


\section{RESULTS}
\label{results_sec}
The algorithms presented have been executed on a computing platform equipped with an 8 core Intel Xeon E 2278 GE, an NVIDIA RTX Quadro 8000 GPU, and 64 GB DDR4 RAM. 

The vehicle position was provided by a localization module based on an Extended Kalman Filter (EKF) using data produced by two GNSS RTK systems with Inertial Measurement Units (IMU) and wheel speed sensors. The perception module exploited the three solid-state LiDAR sensors, one frontal radar, and six cameras mounted on the racecar in order to estimate the obstacles and opponent position. Further details on the whole autonomous software stack will be presented in a future work.

Both the Frenét-Frame-based planner and the motion forecasting module run at a frequency of 20Hz. The planner uses a time horizon of 3s, a sampling time of $\Delta t$ = 50ms, and lateral node sampling of 0.5m. The hard lateral safe distance is set to 3m, and the additional soft margin to 1.5m.
The same $\Delta t$ is used in the MPC with a prediction horizon of $T = 50$ resulting in a time horizon of 2.5s. However, the MPC is executed at a frequency of 100Hz.

Experimental results have been produced in different scenarios at Lucas Oil Raceway (LOR), IMS and LVMS.
\begin{figure}[htb!]
	\centering	
	\includegraphics[width=1\columnwidth]{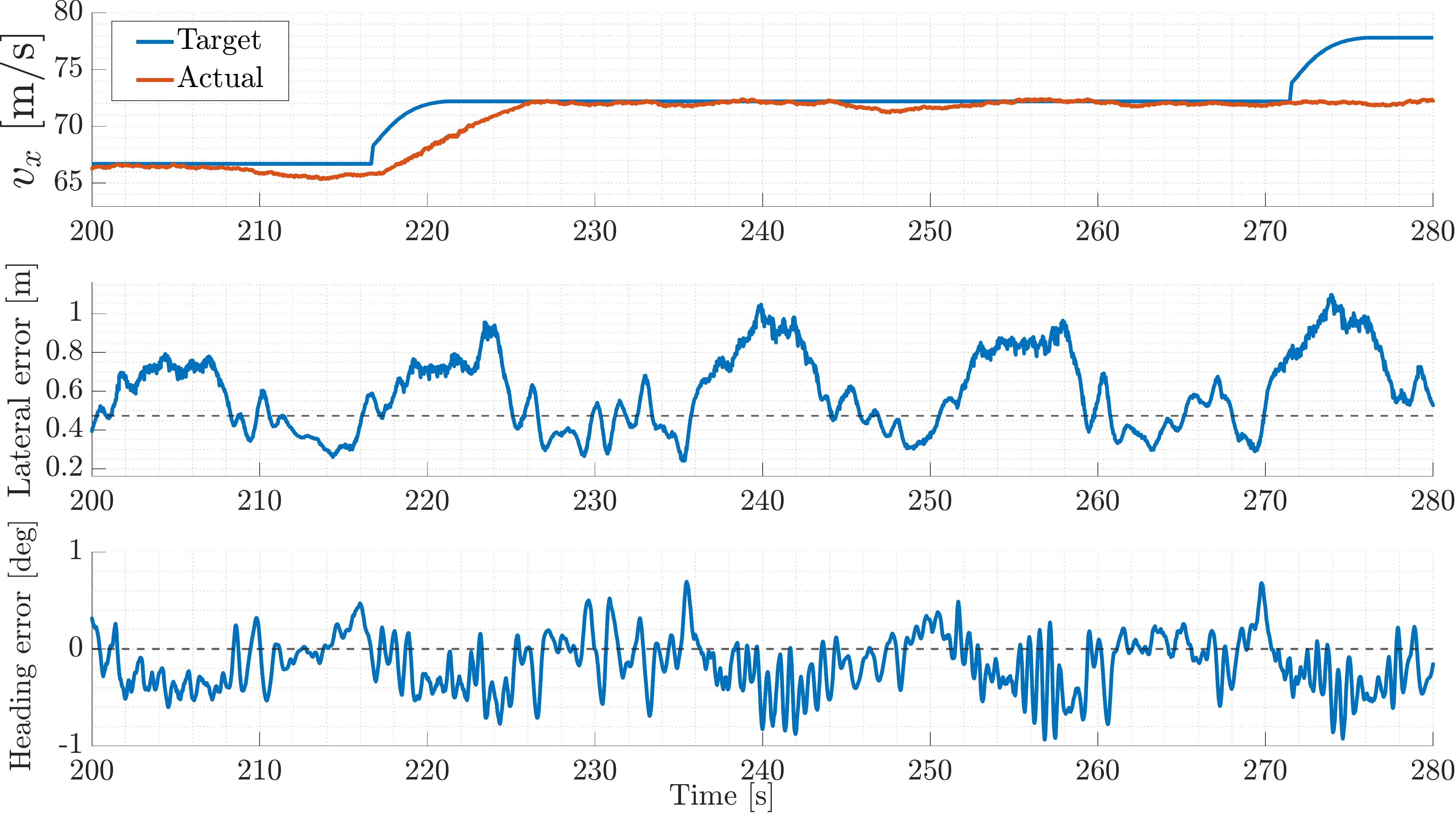}
    \caption{Experimental results for path tracking during high speed laps. See \url{https://youtu.be/ERTffn3IpIs?t=2013} for the time trial laps at LVMS.}
	\label{fig:lvms_hotlap}
\end{figure}

\subsection{High Speed Laps}

\begin{figure}[htb!]
	\centering	
	\includegraphics[width=1\columnwidth]{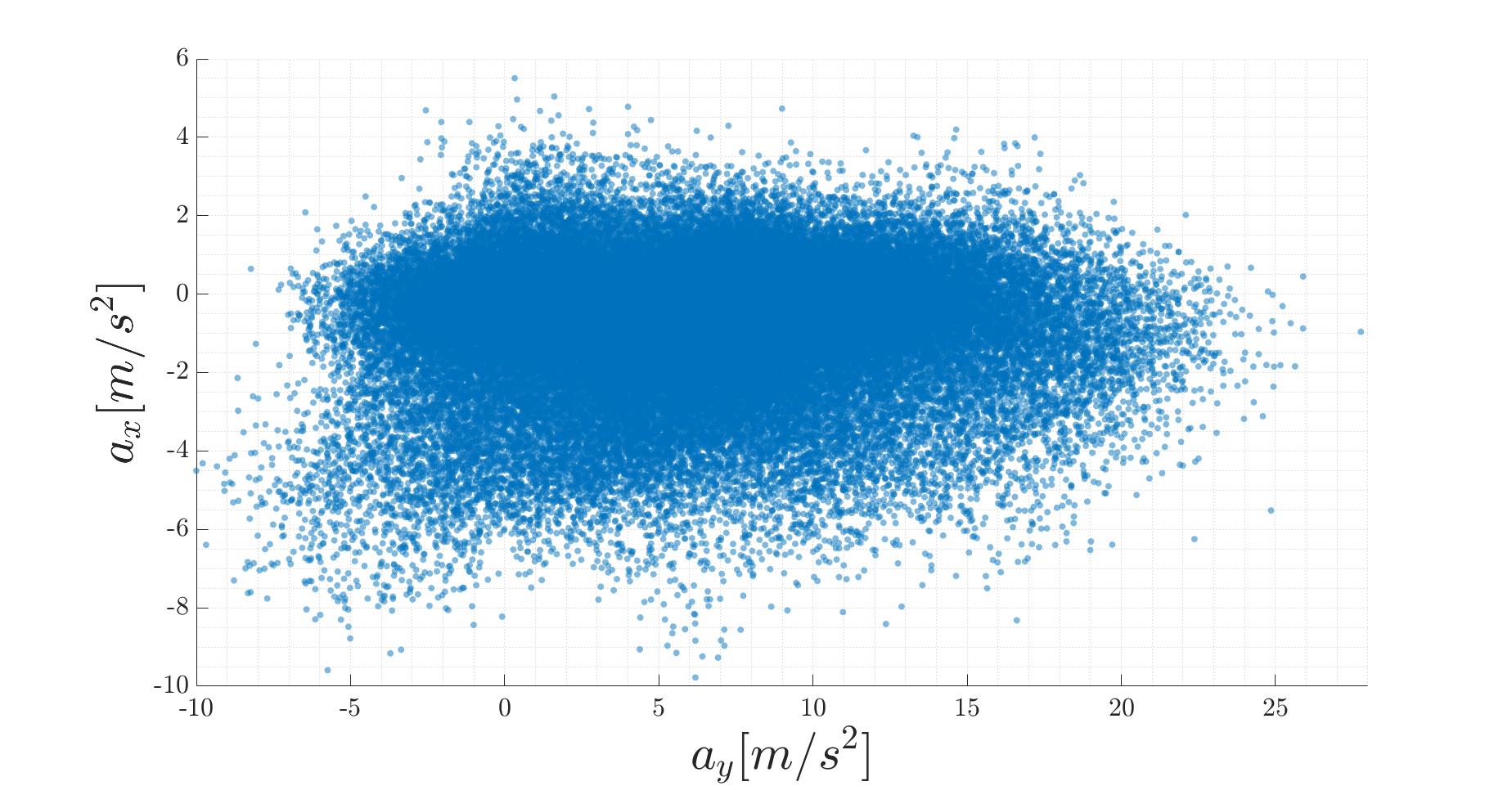}
    \caption{The g-g diagram of the fastest lap at LVMS.}
	\label{fig:gg_diagram}
\end{figure}

The capability of the designed MPC to control the vehicle at high speeds has been tested during the time trial part of the IAC events. At the IMS track, TII-ER achieved the fastest time reaching an average speed of 62.5 m/s over a lap. The tracking performance at the LVMS is shown in Figure \ref{fig:lvms_hotlap}, where a top speed of 75.5 m/s has been reached. The maximum lateral error is 1m, whereas the RMS value is 0.5m. The heading error is maintained between 0.7deg and -1.0deg. A limited heading error despite a not negligible lateral error is the expected effect of the strategy applied to the MPC regularization terms explained in \ref{mpc_sec}. The positive lateral error could be related to a not accurate force offset $Sh_{v_i}$ used in \eqref{eq:model:pacejka}, which should be investigated using the experimental data. Figure \ref{fig:gg_diagram} presents the g-g diagram, showing that the racecar reached lateral accelerations up to 25 m/s\textsuperscript{2}.

\begin{figure}[htb!]
	\centering	
	\includegraphics[width=1\columnwidth]{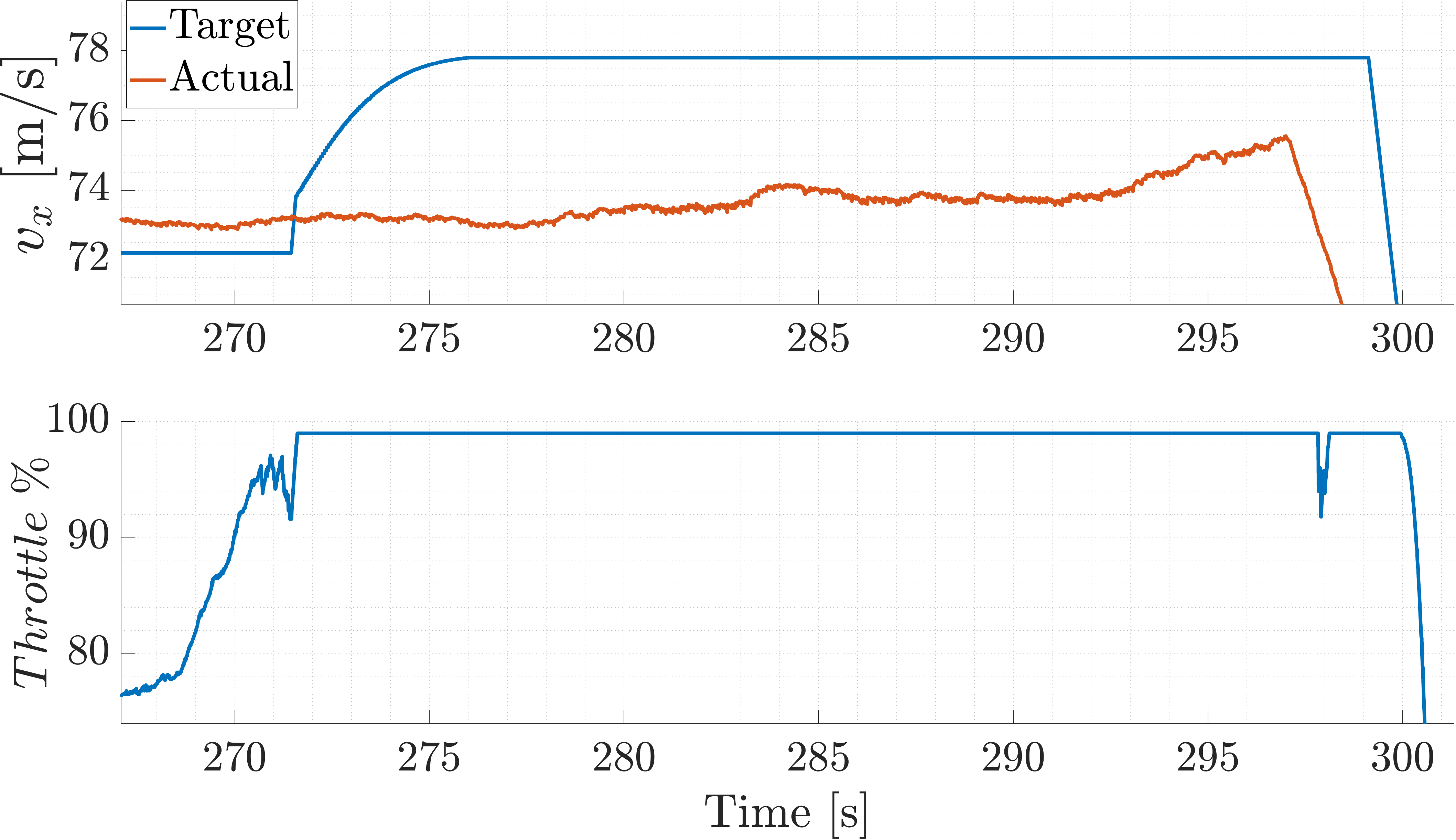}
    \caption{From Figure \ref{fig:lvms_hotlap}: powertrain issue during the high speed laps. A top speed of 75.5 m/s has been reached instead of the target speed of 77.7 m/s.}
	\label{fig:throttle_fail}
\end{figure}

In both events, the top speed has been limited by hardware and engine malfunctions. In particular, Figure \ref{fig:throttle_fail} shows the case in which the research vehicle was not able to reach the target speed despite a fully saturated throttle command due to a detached cable in the powertrain wiring.

\subsection{Static Obstacle Avoidance}
\begin{figure}[htb!]
	\centering	
	\includegraphics[width=1\columnwidth]{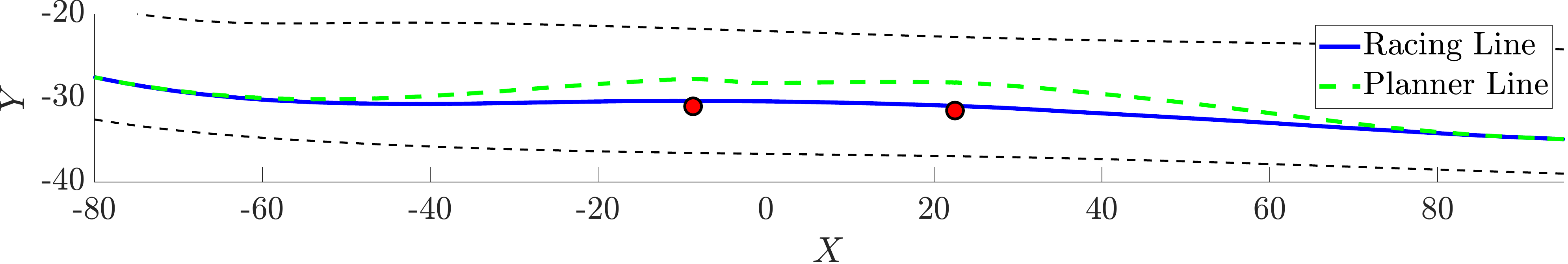}
   \caption{Static obstacles test at LOR. See \url{https://youtu.be/LIzb-_8vrI8} for a video of the test.}
	\label{fig:pilons_lor}
\end{figure}

Figure \ref{fig:pilons_lor} depicts a scenario in which static obstacles have been added to the LOR track. The AV-21 racecar was able to safely avoid the obstacles at a velocity of 34 m/s. The safe sensor range for the LiDAR-based detection was set at 60m. Thus, the planner received the obstacle position 1.7s before the potential collision.
\begin{figure*}[]
	\centering	
	\includegraphics[width=1.8\columnwidth]{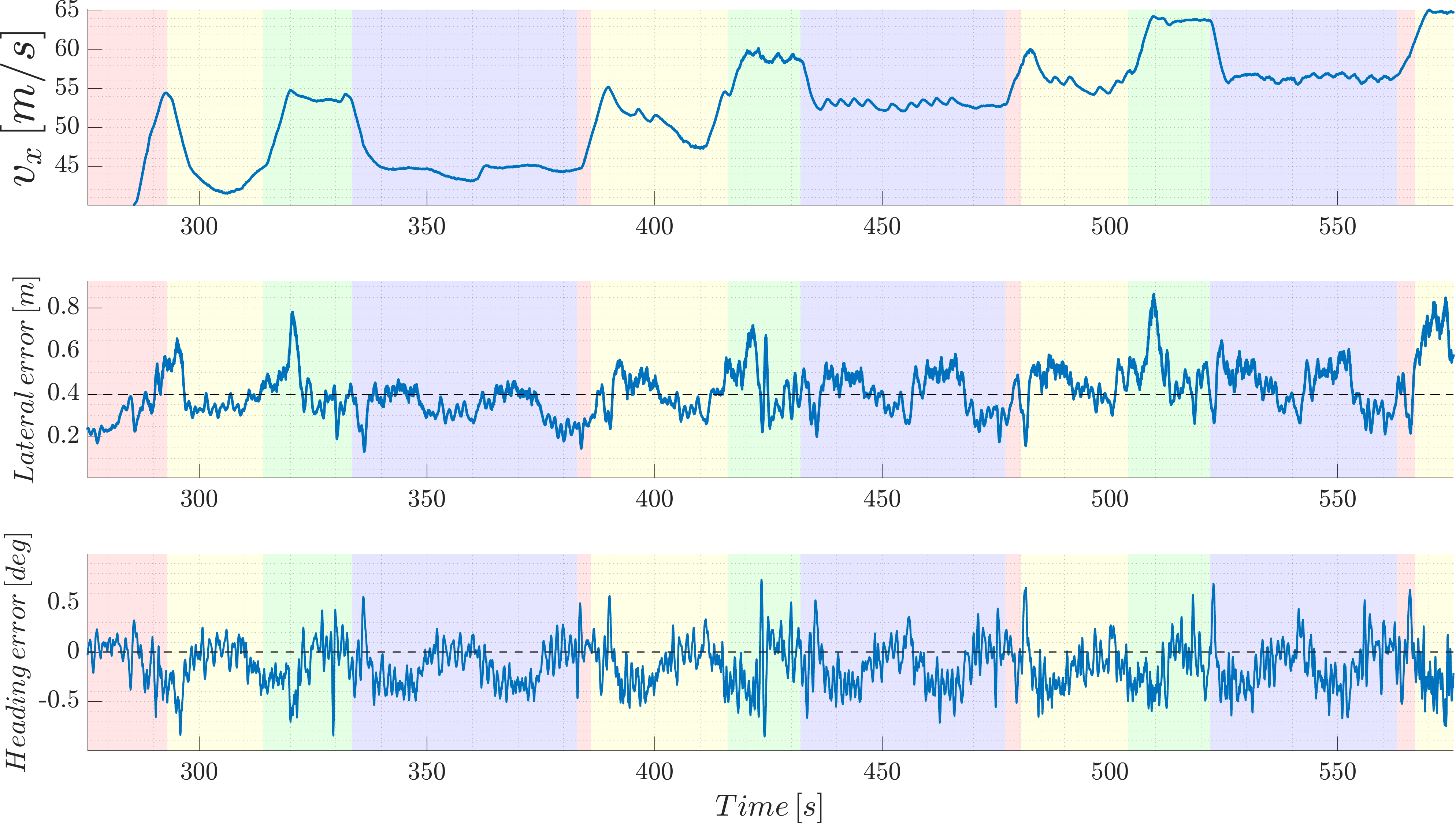}
    \caption{Experimental results for path tracking during the head-to-head race. The four steps are depicted with different colors. Closing the gap (red). Following mode (yellow). Overtake (green). Defending (blue). See \url{https://youtu.be/ERTffn3IpIs?t=10469} for a video of the complete match.}
	\label{fig:lvms_perf}
\end{figure*}
\begin{figure*}[]
	\centering	
	\includegraphics[width=1.9\columnwidth]{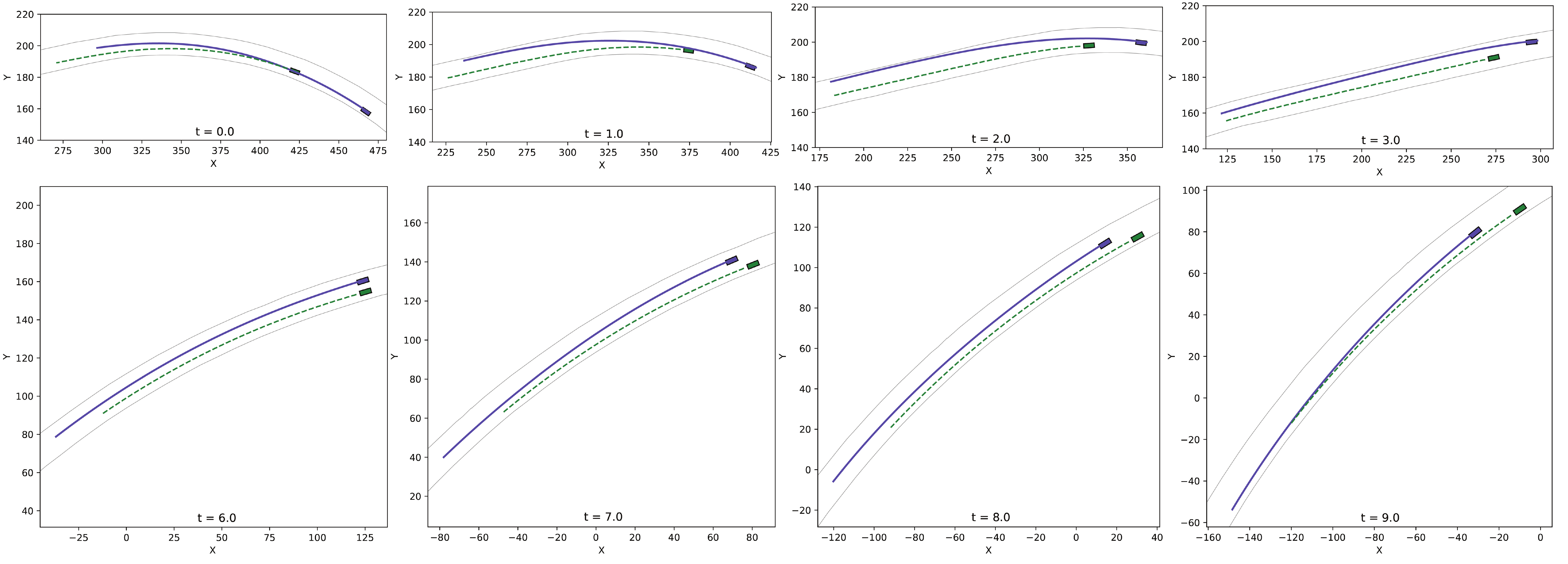}
    \caption{Frame sequence during the overtake at 63 $m/s$. The solid blue line is the planner trajectory. The dashed green line is the motion prediction of the opponent.}
	\label{fig:overtake_frames}
\end{figure*}
\subsection{Head-to-Head Racing}

A passing competition was held at LVMS where the racecars had to perform overtakes at increasingly higher velocities respecting the race format composed of four steps. The attacker should first reduce the gap from the defender, keep a longitudinal safety distance and overtake once reaching a passing zone. Then, the roles can be switched. If the new attacker succeeds in the four steps, a new round at higher speed is started. Figure \ref{fig:lvms_perf} shows the performance of our solution during the last four rounds of the semifinal. Similar results of the time trial have been obtained with a higher lateral and heading error during the initial portion of the overtaking maneuvers. Figure \ref{fig:overtake_frames} shows a frame sequence of the overtake at the highest speed achieved. It should be mentioned that the experimental data in the head-to-head scenario ends at a top speed of 63 m/s due to a wrong hard brake command triggered by a module separated from the motion planner and controller, causing the TII-ER vehicle to collide with the track borders.


\section{CONCLUSIONS}
\label{conclusion_sec}
A multi-body model of the racecar has been implemented in simulation and used to examine and identify non negligible dynamics prior to the tests on track. This approach combined with a higher weight on the steering rate of change term demonstrated to be a successful strategy in making the controller robust enough at velocities of 75.5 m/s and accelerations of up to 25 m/s\textsuperscript{2}, which were never explored before the final racing events.
The planner has been capable of generating a path and velocity profile in order to follow the opponent, maintaining a defined distance, and producing a safe trajectory for active overtakes at speeds up to 63 m/s.
Experimental data gathered during the tests will be used to improve the model identification and regularization terms, aiming to explore the dynamics and tire friction limit of the vehicle, as well as reduce the path tracking error. 
More challenging scenarios, such as racing in complex road courses and competing against multiple agents, will be explored in future applications.
Further research will focus on new approaches for online trajectory generation to accomplish more aggressive maneuvers to adapt to racing conditions while keeping into consideration safety and computational limitations.
\addtolength{\textheight}{0cm}   



\section*{ACKNOWLEDGMENT}
\label{ack_sec}
The authors would like to thank all the members of the TII EuroRacing{\footnote{\href{https://www.tiieuroracing.com/}{https://www.tiieuroracing.com/}} team. We would also like to thank MegaRide{\footnote{\href{https://www.megaride.eu/}{https://www.megaride.eu/}}for the support in the tire model identification, in particular Lorenzo Mosconi, Ph.D. Student from University of Naples Federico II. Thanks to Claytex{\footnote{\href{https://www.claytex.com/}{https://www.claytex.com/}} for providing the VeSyMA Motorsports Library.


\end{document}